# Intelligent Parsing: An Automated Parsing Framework for Extracting Design Semantics from E-commerce Creatives

Guandong Li[1*], Xian Yang[1]

1.*Suning, Xuanwu, Nanjing, 210042, Jiangsu, China.

*Corresponding author(s). E-mail(s): leeguandon@gmail.com

*Abstract*: In the industrial e-commerce landscape, creative designs such as banners and posters are ubiquitous. Extracting structured semantic information from creative e-commerce design materials (manuscripts crafted by designers) to obtain design semantics represents a core challenge in the realm of intelligent design. In this paper, we propose a comprehensive automated framework for intelligently parsing creative materials. This framework comprises material recognition, preprocess, smartname, and label layers. The material recognition layer consolidates various detection and recognition interfaces, covering business aspects including detection of auxiliary areas within creative materials and layer-level detection, alongside label identification. Algorithmically, it encompasses a variety of coarse-to-fine methods such as Cascade RCNN, GFL, and other models. The preprocess layer involves filtering creative layers and grading creative materials. The smartname layer achieves intelligent naming for creative materials, while the label layer covers multi-level tagging for creative materials, enabling tagging at different hierarchical levels. Intelligent parsing constitutes a complete parsing framework that significantly aids downstream processes such as intelligent creation, creative optimization, and material library construction. Within the practical business applications at Suning, it markedly enhances the exposure, circulation, and click-through rates of creative materials, expediting the closed-loop production of creative materials and yielding substantial benefits.

Key words: Intelligent parsing, Design semantic, Classification, Detection, Recognition, Intelligent design

1.Introduction

In the industrial e-commerce landscape, creative designs such as banners and posters are ubiquitous [1,2]. Extracting design semantics from creative e-commerce graphic design materials is one of the core challenges in the field of intelligent design. These creative e-commerce materials typically stem from designers' works, such as drafts created using digital tools like Figma, PSD, Sketch [3], which eventually give rise to a series of e-commerce images. Traditional creative designs often involve modifications based on these designer works. However, when the creative loses its relevance for deployment, the original creative material is discarded, leading to significant wastage of creative assets.

In recent years, with the development of intelligent creative design and the establishment of digital marketing platforms, there has been a unified medium for the usage and circulation of creative design works, from their initial design phase to the eventual generation of programmatic creatives. Designers' design drafts undergo intelligent parsing, and the semantic information derived from this parsing is commonly utilized for intelligent creation, such as products like Smartbanner [4]. Smartbanner, utilizing template materials parsed intelligently through a planner, actioner, tuner, and generator framework, achieves an intelligent banner design framework that balances creative freedom and design rules. The additional creative tags attached by the intelligent parsing's labeling system are used for continuous creative optimization, for instance, creative optimization [5]. Dynamic creative optimization converts the tags of creative materials into feature dimensions, thus realizing a creative precision-ranking framework in the first phase. Materials parsed intelligently are also used for constructing material libraries, enriching material creativity, and enabling the upgrade and optimization of intelligent marketing platforms through tagged management updates. These advancements facilitate the utilization and circulation of creative material designs. Therefore, an automated and highly intelligent creative parsing framework yields significant benefits downstream in the entire creative design process, as illustrated in Fig.1 with our applied intelligent parsing framework.

The existing intelligent parsing frameworks primarily focus on automatic code generation [6,7]. However, there is a significant gap between the design drafts (similar to PSD, Sketch, and other digital tools) and high-quality products (code and its visual representation) during the process of automatic code generation. Therefore, it is necessary to parse the design drafts in order to generate structured outputs that can be further transformed into code-based visual representations. Unlike our creative intelligent parsing approach, intelligent parsing is more diverse and comprehensive. It involves processing different types of design drafts and achieving richer and structured outputs through material recognition layer, preprocessing layer, smartname layer, and label layer. It requires incorporating style tags in the design and various tags and structural components at the creative design layer level.

In general, intelligent parsing includes: 1) Material recognition layer, which is based on the creative design information and divides it into three levels: global level, local level, and layer level. The information at each level is further classified into composite information and native information. The core capability of the material recognition layer is to identify and detect composite information through methods such as product text area detection, detailed text area detection, creative element center point detection, transfer learning classification, label recognition classification, cascadercnn [8], gfl [9] multi-level detection methods, and single/multi-label classification and recognition [10,11]. This results in obtaining information at the three levels. 2) Preprocessing layer, which mainly achieves recognition of creative material types. Intelligent parsing supports multiple types of creative material parsing, layer filtering, removal of unusable and non-standard design layers, and material grading, dividing creative materials into high-quality/editable/unusable categories for efficient usage. 3) Smartname layer, which is the core parsing layer. It accurately obtains prior design result information by renaming each level and identifying the position of each layer. 4) Label layer, which is divided into creative material layer and creative layer level. Different labels are designed for different creatives, and the creatives are classified and structured according to their style, category, and scene, for future use.

Our contributions are as follows:

1.We propose an effective and highly intelligent parsing solution in the field of creative design analysis. This solution achieves automated retrieval of design semantics, providing structured information output for downstream design optimization. Most intelligent parsing frameworks only cover a partial aspect of structural information, but intelligent parsing fully implements the entire creative parsing process.

2.In the smartname module, we propose a comprehensive four-step solution consisting of text preprocessing, first-stage processing, second-stage processing, and post-processing to address the challenges of layer renaming and position adjustment. The identification and localization of design elements in the original designer's drafts can be difficult and typically do not conform to specific standards. Therefore, a comprehensive consideration, continuous filtering, and logical selection are required to obtain satisfactory results.

3.In the overall architectural design, we abstract the material recognition layer to classify all information in the creative design draft into three categories: global, local, and layer levels. Based on this classification, we define native and composite information, ensuring their integration throughout the intelligent parsing architecture.

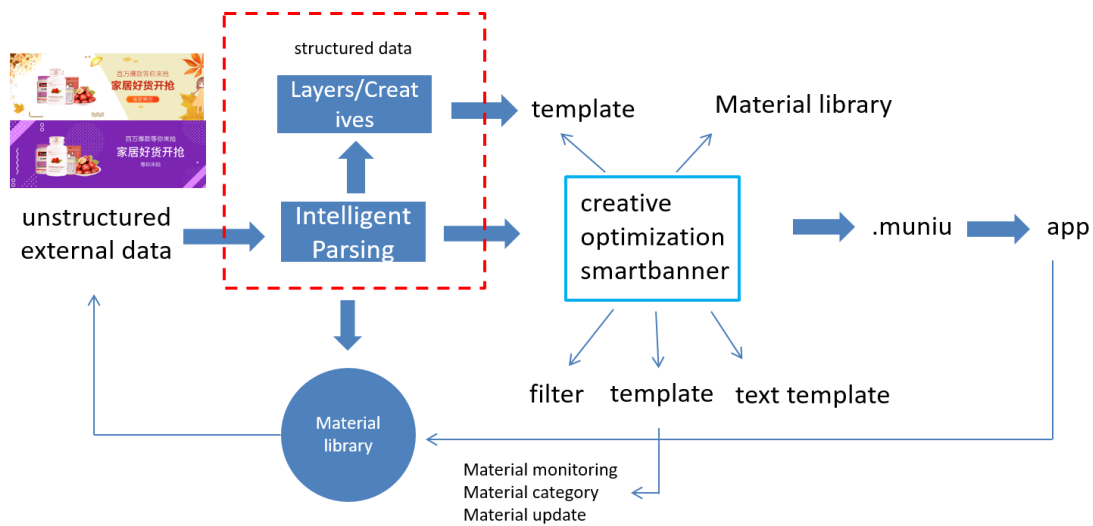

Fig.1 Downstream application framework for intelligent parsing

2.Related works

2.1 Creative Design

A smart banner design framework called Smartbanner has been proposed to achieve a balance between creative freedom and design rules. Smartbanner consists of a planner, an executor, a fine-tuner, and a generator. Yang et al. [12] designed a system for automatically generating digital magazine covers by summarizing a set of theme-relevant templates and introducing a computational framework containing key layout design elements. These works are typically rule-based and have relatively low intelligence. Vinci [13] proposed a design space to describe design elements in advertising posters and introduced design sequences to provide feedback on the design decisions made by human designers when creating posters. In the industry, Alibaba's Luban [14] and JD.com's Linglong system are also mentioned.

2.2 Intelligent UI code generation

Automated UI code generation has become an attractive topic since the rise of artificial intelligence. Early intelligent automation research was primarily aimed at replacing template-based UI design, where users had to spend time searching for suitable materials to assemble their designs. Batuhan et al. [15] utilized hand-drawn images to recognize and generate basic buttons, text, images, and other components. Works like sketch2code[16] leveraged more detailed design sketches and achieved automated generation of UI structures. Similarly, Pix2code[6] generated textual descriptions from photographs in a similar manner. It achieved high accuracy by taking actual UI screenshots as input.

2.3 UI design check

With the popularization of automated UI generation technology, it is necessary to evaluate the quality of the generated UI and address issues such as missing elements or component overlaps caused by hardware or software compatibility. OwlEye [17] detects GUI display issues based on deep learning methods and locates detailed areas of the problem. LabelDroid [18] focuses on image-based buttons and achieves highly accurate prediction of labels through learning from a large-scale commercial applications from Google Play. FSMdroid [19] dynamically analyzes GUI applications using MCMC sampling methods and detects defects that exist on less frequently accessed paths.

3.Methods

Parse creative design drafts from different sources (internal/external site) and in different styles (banner/poster/product main image/detail image) into a custom structured data format, selectively store high-quality materials (template/image) into the database, providing massive data support for subsequent intelligent creation (Smartbanner). It can also serve as tagged information for subsequent creative selection, continuously optimizing creative selection and creative materials.

The overall process of intelligent parsing is shown in Fig 2, and the specific structure of a certain block is shown in Table 1.

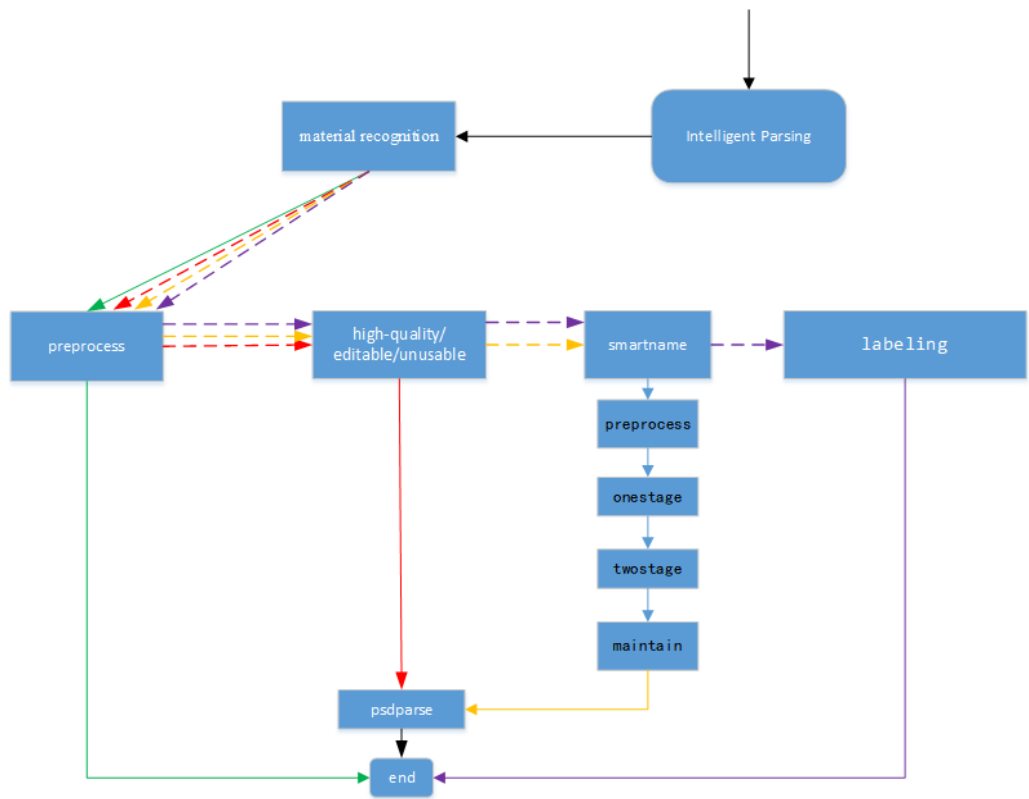

Fig 2. Intelligent parsing process diagram

Table 1. Specific module and module implementation function table.

|  | Methods | | Specific implementation module |
|---|---|---|---|
| Input | Creatives(PSD/Sketch…) | | |
| Layer reading layer | Psd_tools | | Layer object collection and images |
| Material recognition | Detection,Classification | | Psd type recognition, material recognition, layer label recognition, region detection |
| Preprocessing | Psd filter | Classification,Detection,Expert rules | Psd type recognition, banner filtering, layout detection, image deduplication, redundant layer detection |

|  | layer preprocess | Image retrieval, detection, expert rules | Deleting transparent/useless layers, layer effect conversion, competitor logo detection |
|---|---|---|---|
| Smartname | Process | Classification, OCR, expert rules | Text/image correction, text recognition, text bounding box, text naming |
| | Onestage | | Goods/background recognition, text layout correction |
| | Twostage | | Layer correction recognition |
| | Maintain | | Mcde separation and text layout correction |
| Label | Classification model, location tag determination, primary color extraction, color transfer | | |
| Structured Output Layer | | | |

3.1 Material recognition layer

Creative design materials are often presented in PSD format. Intelligent parsing explores how to obtain structured semantic information from the PSD format data to assist creative design and optimization. PSD organizes data in a layer structure, and the PSD format preserves the layered structure of images. Each layer can contain different elements, such as images, text, adjustment layers, effects, etc. This allows non-destructive editing, as individual layers can be modified or hidden without affecting other layers.

The information contained in the layers of the PSD structure is divided into three levels: global level, local level, and layer level. Under each level, the information is further classified into two categories: composite information and native information. The composite information refers to the design information obtained through detection and classification methods, including custom product areas, text areas, etc. The native information refers to the information inherent to the PSD layers themselves, including layer styles, text styles, coordinates, opacity, and other data attribute information.

3.1.1 product text area region detection

Banners and other e-commerce creative materials often have separate areas for the main product and text. The product area represents the merchandise that the creative intends to showcase, while the text serves as the focal point and promotional slogan. We have utilized Cascade R-CNN to train a model for detecting the product and text areas.

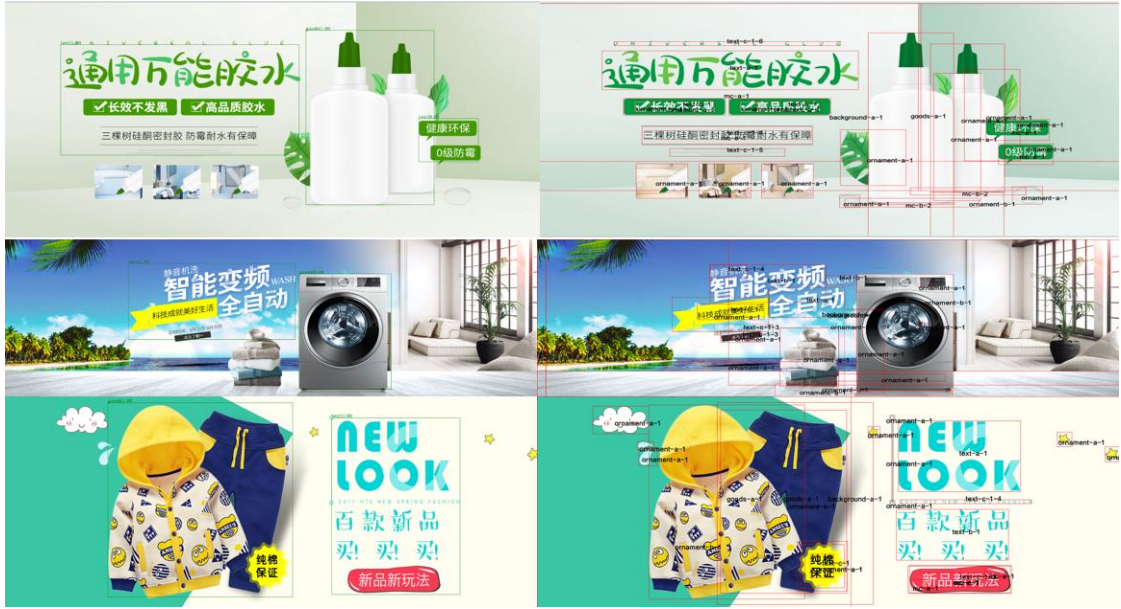

Fig 3 product text area region detection examples

After obtaining the corresponding regions, we need to filter out the target layers from these regions. Since the composite information of the global layer is detected on the PSD image, it is necessary to map the product and text to specific layers. All layers with an intersection-over-union (IOU) greater than 0.8 with the product area and within the detection region of products are usually considered as product layers, and the same applies to text layers. However, the type layer of the text is removed in the product area.

3.1.2 Fine-grained detection of text areas.

Using Cascade R-CNN, the text areas are further divided into "horizontal," "vertical," "mask," "maskbox," "wordart," and "shape." After obtaining these regions, we consider filtering out the layers with high IOU within each region.

3.1.3 Mcb/mccde center point detection

The MC belongs to the decorative layer of products and text, and usually requires determining the center point coordinates of the layer. Using GFL_R50 model.

3.1.4 Transfer Learning Classification

This module supports the layer classification module in the smartname module and the layer filtering module in the preprocessing. The transfer learning model in this module selects vgg16 as the base model, and only fine-tuning the last classification layer is needed.

3.1.5 Label Recognition Model

This module supports the tag recognition algorithm for the tag layer. At the algorithm level, it includes single-label recognition and multi-label recognition. Single-label classification uses the res2net algorithm to transform the input channels, enabling it to input four-channel transparent images. Multi-label classification uses the ASL multi-label classification algorithm. At the business logic level, it includes tags for the entire creative and specific layers in the creative; Tags at the creative level include: style, category, background, and scene. Tags at the creative layer level include: entity, entity subcategory I, style, category, tag form; More tags will be described in detail in the tag layer.

3.2 Preprocessing Layer

The preprocessing layer primarily involves automatically identifying the type of input creativity and further subdividing it. First, the creative materials are filtered to remove and clean up layers that do not meet the requirements. The creative materials are then scored, and high-quality creative materials undergo further processing to obtain design semantic information.

3.2.1 Creative Material Type Recognition

The input image of "intelligent parsing" is classified into five categories, including banner, poster, product main image, detail image, and others. Automatic recognition is performed on these five types of images.

3.2.2 Layer Filtering

1. Transparent layer deletion

Determine whether the layer is a transparent layer, and if so, delete it.

2. Layer effect migration

Migrate some required layer effects.

3. Advertisement layer filtering

The image retrieval module uses senet18 to extract image features as queries and designs an advertisement layer library. The features are extracted in advance. Generally, advertising marks will appear in the first or second layer of the layer. The images in the layer are used as queries, and the cosine similarity is calculated between the already extracted feature advertising marks in the preset image library to filter out advertising mark layers.

4. Similar image retrieval

In the PSD creative material designed by the designer, there may be multiple duplicate images, so we use DHASH for quick deduplication when filtering out duplications in the creative materials.

5. Logo Detection and Recognition

Logos are usually detected and screened in creative materials, as logo information needs to be preserved in subsequent applications, including intelligent creation and optimization. Therefore, we use two methods in logo detection:

a. Logo detection + recognition

b. Calculate the cosine similarity with the preset logo image library to determine the logo to which the creativity belongs

6. Rewrite Text Image Layer Properties

In creative materials, text materials designed by designers are usually in the form of type layers, which are editable text layers. However, some designs use large image layers instead of text layers, which may simplify the design process but make it impossible to parse text information and properties for downstream creation and optimization tasks. Therefore, it is necessary to use OCR technology in this layer to recognize the text in the images and fill in the attribute information.

a. We trained a classifier for text and image layers using res2net50. If a layer is not a type layer but is classified as a text layer, it is most likely a text image layer.

b. Perform OCR detection on the text image layers using the DBNet+CRNN OCR architecture. However, direct detection and recognition of text within the layer have poor results. Therefore, the image is usually padded and expanded to 736x736 to obtain the text content and size of the layer.

c. The identified text should be within the designated text area of the layer.

### 3.2.3 Creative Material Grading

In the intelligent parsing framework, we grade creative materials, where high-quality materials can be automatically extracted for design information and form structured data. We divide creative materials into three levels: high-quality, editable, and unusable. Each level is processed using corresponding methods.

1).Element Quality Assessment

In PSD files, there may be a large number of repetitive line segments or they may be composed of numerous line segments. We trained a lines classification model using transfer learning architecture. This model is integrated into the material recognition layer to predict the presence of line frames. Additionally, we utilize DHash to detect the repetitiveness of layers within a group. If the repetition ratio or line frame ratio is excessively high, we lower its layer level.

2). Canvas Assessment

During the intelligent parsing of PSD files, we have observed that in scenarios with poor design practices, many canvases are not properly cropped. When the canvas extends significantly beyond the visible area and contains a large number of layers placed outside the canvas, it often results in a reduction of the creative material's layer hierarchy.

3).Product Recognition Assessment

The conformity of products is an important aspect in determining the quality of creative materials. We trained a product recognition model using a transfer learning approach within the material recognition layer, resulting in a binary classification model for products. Additionally, we combine the product recognition area to assess the quality of the identified products.

4). Usable text Area

We perform detection on the text areas obtained from the material recognition layer. The training process utilizes our in-house developed Tpclas model. If a layer represents an unusable text area, it will be filtered out.

5).Comprehensive Hierarchy

a. Assessment a. Determine if there are multiple product layers and if there are any products present. The product layers are obtained using the product recognition model.

b. Calculate the IOU (Intersection over Union) ratio between the product layers and the copy layers in the global PSD file.

c. Assess the composition of grouped elements.

d. Check for the presence of complex copy areas.

Based on the above information, a hierarchical classification is assigned to the creative material.

### 3.3 Smartname layer

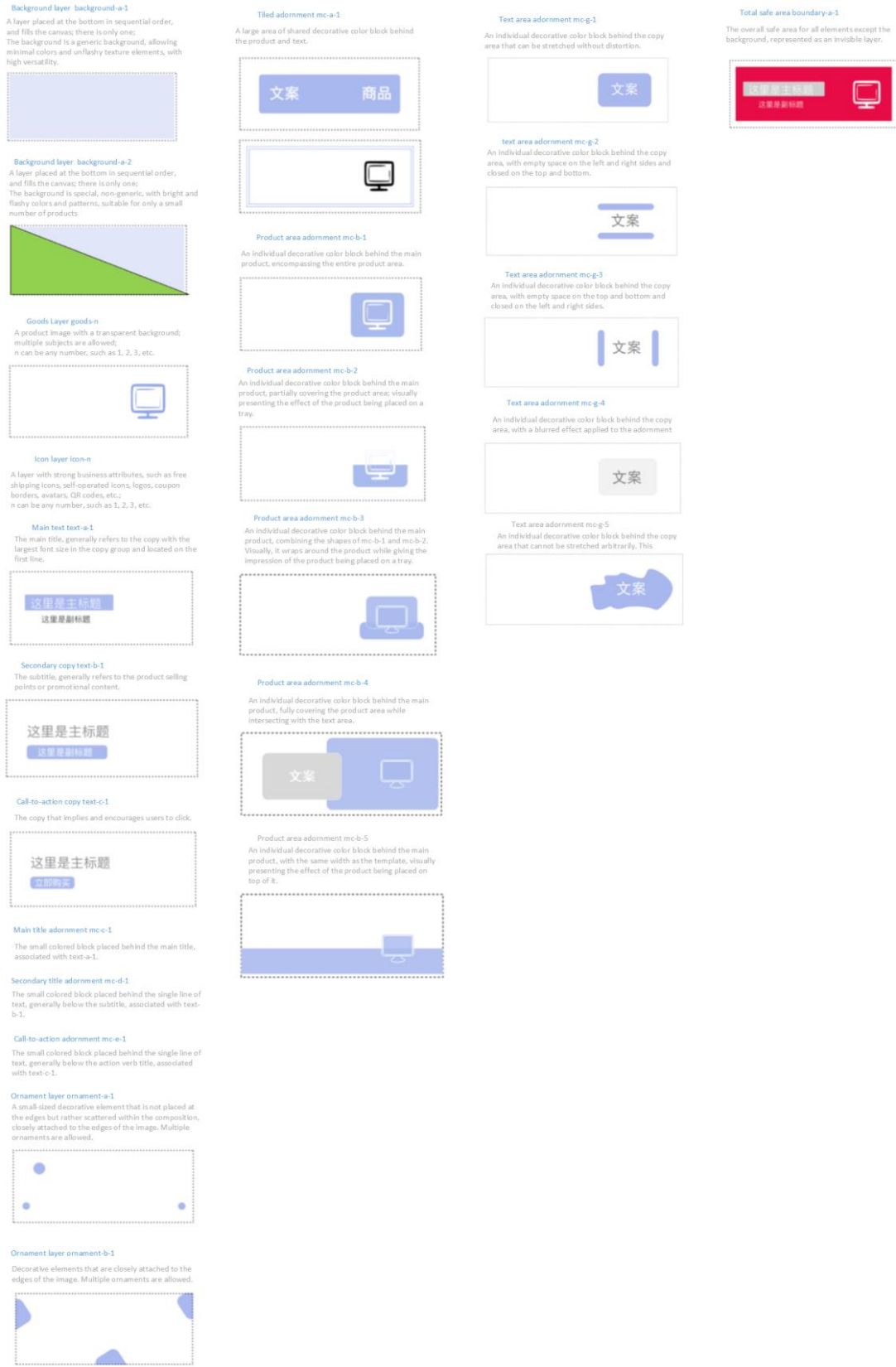

Fig 4 smartname-specific category names and classification basis chart

The smartname layer is the core layer of intelligent parsing, which typically involves labeling and categorizing creative materials. In this process, the layer names are finely divided. The specific categories are shown in Fig 4.

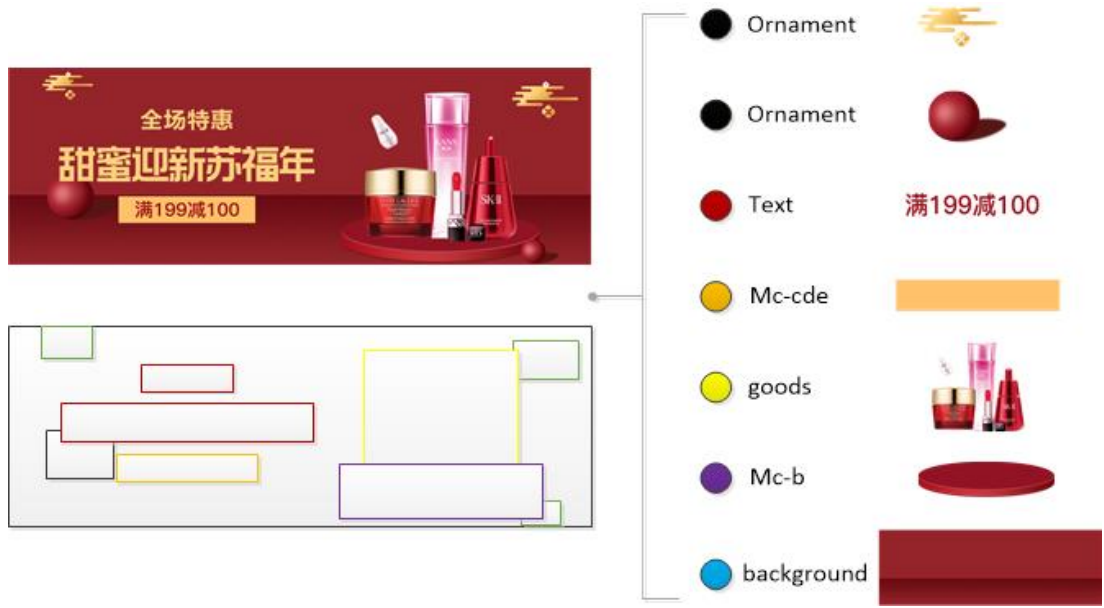

Fig5 The display shows the major categories of layer names. In fact, for the ornament layer, we have subcategories such as ornament-a-1 (border decoration layer) and ornament-b-1 (non-border decoration layer). As for the mask layer, different types of masks are generated by matching with different copy layers. For example, matching with the main copy is referred to as mc-c, matching with the secondary copy is referred to as mc-d, and matching with action verbs is referred to as mc-e, and so on.

In the algorithm recognition module, we use a transfer learning recognition method from the material recognition layer and employ a multi-level recognition approach. Through a coarse-to-fine hierarchical recognition process, the layer names are obtained by continuously narrowing down the solution space. As shown in Fig 6.

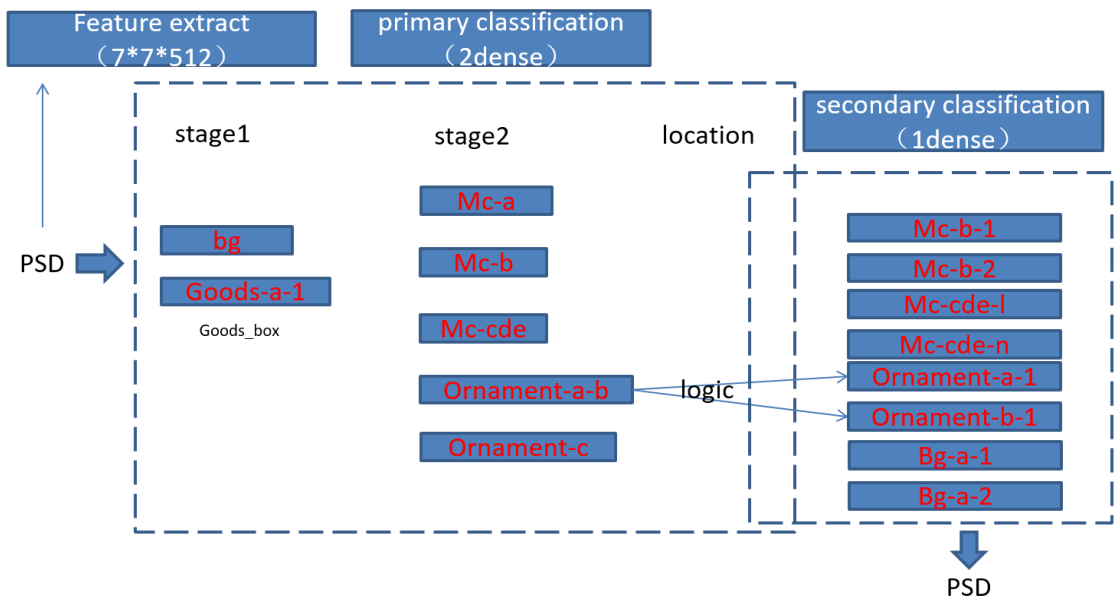

Fig 6 The technical architecture diagram of smartname

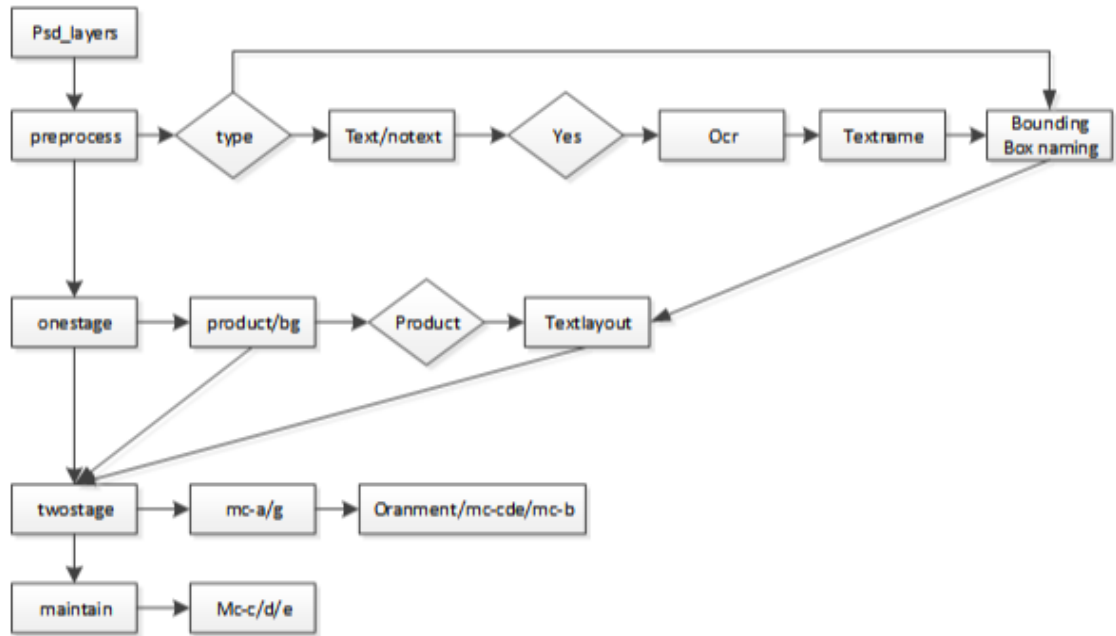

Fig 7 Technical Workflow Diagram of Smartname

This system divides smartname into four steps. The specific algorithm flowchart is shown in Fig 7.

3.3.1 Text Preprocessing

In the parsing of creative materials, the text layer can be predicted in advance through the prior knowledge of the type layer in the PSD file. Therefore, as a category prior, the text layer is usually parsed as the first step of preprocessing. However, the text layer is also complex and requires further processing.

a. Bounding Box

In material design, text is usually not in one layer but distributed across multiple layers. Even after obtaining the copy area, it is still necessary to precisely classify the text layers by creating bounding boxes around individual characters and annotating them as defined layers. Based on the angle of each box and the merged boxes within a line, the coordinates of the connected rectangular boxes inside each box are calculated by setting angle interval thresholds, vertical interval thresholds for grouping by row, differences in text line heights for merged boxes, and thresholds for box-to-box spacing to create bounding boxes.

b. Post-processing of Bounding Boxes

The average font size of all individual boxed characters is taken after creating the bounding boxes, and any missing boxes that may have been lost during the bounding process are restored.

c. Sorting by Box Height Sorting

Based on font size may result in errors, therefore sorting is done based on the height of the bounding boxes.

d. Preliminary Naming of Text Layers

Based on the naming rules of text layers in this system, the topmost text box in the entire creative material is typically named text-a-1, followed by text-b-1, and finally text-c-1. If there are additional text layers, they are uniformly named text-c-1, usually representing call-to-action text. Note that at this stage, we have obtained the preliminary copy layers and their corresponding coordinates. However, the core purpose of text preprocessing is to perform tasks such as bounding box creation on text layers, transforming them into true layers instead of separate text elements. In the first-stage prediction, more complex alignment steps will be taken for further determination and validation.

3.3.2 One-stage Prediction

Background and Product Identification Background is relatively easy to identify, while products have a high priority throughout the creative material. The copy area can be obtained through text preprocessing, and usually the positions and names of other elements need to be determined based on the product and copy area. Although the model has identified the product layer during the grading of creative materials, the product layer is actually quite complex, and there may be confusion with other decorative layers, requiring some post-processing logic verification and judgment, which is a validation logic model.

1.Background Identification

Usually in PSD, the layer with sequence 1 is the background layer, but the background layer is often easily confused with ornament-a/b. Ornament-a/b are usually large and scattered decorative layers. When preprocessing the layers, we fill the transparent background of the layer with a fixed pixel value. Therefore, when identifying the background, we first determine whether the fixed areas (usually four corners and center) are filled with the fixed pixel value. If not, it is identified as the background. This system has two classifications for the background, and uses a transfer learning-based binary classification method to classify the layers.

2.Product Identification

The product layer determined by the model is known, and it needs to be verified with the product copy area detection model from the material recognition layer.

a. Validation Logic Model:

1).For the layers identified as products by the model, calculate the Intersection over Union (IoU) with the product copy area detected by the material layer. If it meets the requirements, retain the layer. The product copy area refers to the detected area of the product copy in the material layer.

2).If no product layer is found but a unique ornament layer is found, define the ornament layer as a product. However, since some creatives may not have a product (these creatives are defined as editable during creative material grading, but there may still be some missed detections), the determination of a product needs to use a label recognition model to determine if it is a physical entity. Most products are physical entities.

3).During product copy area detection, if there is only one product layer in the detected area, it is considered a product.

4).If there is a product layer during product copy area detection, but the product recognition model does not identify it as a product, consider it as a product layer, giving priority to the detection from the copy area.

5).If no product copy area is detected, the layers identified by the product recognition model are considered product layers.

6).For some products in the product copy area detection stage, if they are not detected by the product recognition model or only a small number of recognitions are detected, there may be missed detections. In this case, divide the product area into four quadrants, and if there is an IoU with both the layer and any of the quadrants, add it as a product layer.

b. Obtaining Coordinates of Product Layers

Retrieve the coordinates of the identified product layers. There may be multiple product layers.

3. Validate the copy area based on the product layers.

Given the product layers, verify the copy area determined by the layer, the product layer, and the copy area detected by the product copy area detection model.

a. Determine the layout based on the product coordinates. The judgment criteria use the recorded coordinates of four points to determine whether they are to the left or right of the designated reference point, in order to determine whether the creative material has a horizontal or vertical layout.

b. Identify the copy areas in different layouts.

c. The determined copy areas may be multiple and distributed in various areas of the creative material. Although some text layers may be recognized, they might only be decorative text rather than actual copy areas. Typically, there is only one copy area in the creative material, which can be identified by sorting and calculating the area.

d. A scoring function is designed based on the area, layout, and the number of text layers within the layout, in order to identify the genuine copy area and the corresponding text layers.

e. Revise the names of the text layers in the copy layer.

### 3.3.3 Two-stage prediction

In the first stage prediction, we obtained precise background, product, and copy layers. Now, in the second stage, we need to determine more refined small element layers such as mc and ornament.

1). Determination of mc-ag Layers

a. The mc-a layer is a decorative layer outside the copy and product areas, and its identification can be achieved by calculating the Intersection over Union (IoU) with the copy and product areas.

b. The mg-g series is a relatively complex type of copy decoration layer. We have designed a classification system for mcg recognition, primarily based on the shape after binarization for identification.

2). Determination of ornament series, mcb, and mccde series

a. By using a two-level classification model based on transfer learning, we can obtain three main categories: mcb, mccde, and ornament.

b. For mcb, further classification is carried out using the transfer learning method. MCB is usually related to product layers and serves as a decorative layer for product layers. However, we need to consider false detections between mca and other layers.

c. For ornament, further classification is carried out using the transfer learning method, but we need to be careful about false detections with mca, mccde, mcb, and other layers.

### 3.3.4 Maintain layer

In the material identification layer, in addition to the detection of the product copy area, we also performed fine-grained detection of the copy area and center point detection for mcb/mccde. However, in the second stage, we did not perform fine-grained recognition for mccde, resulting in a high rate of false detections. Therefore, in the post-processing module, we combine the fine-grained detection of the copy area with the distribution of mccde within the detection area and the coordinates of the mccde center point to further classify mc-c/mc-d/mc-e.

At this stage, through intelligent parsing, we have obtained the hierarchical levels of the creative material as well as the layer names and classifications within the first-level creative material.

3.4 Label layer

Labeling creative materials is a crucial step, and intelligent parsing serves as the upstream process in creative applications. The structured layer information and label data play a significant role in guiding downstream processes such as intelligent synthesis and creative optimization. This module, in combination with material recognition, achieves the design of labels for creative materials. Semantic information extraction in the label layer is divided into creative material hierarchy and layer hierarchy. Within a series of labels, some can be obtained through parsing the format of the creative material itself, such as the psd format, while others require a label system model. In terms of the model, the label layer includes single-label recognition and multi-label recognition, based on the self-developed Tpclas classification detection framework, with inputs being four-channel transparent images based on the res2net model.

3.4.1 Creative Material Hierarchy

Labels In the basic labels of the creative material hierarchy, Viewbox/Layoutcount/ColorMode/Size can be parsed and obtained, while the remaining labels require further model parsing:

a. Color: primary color extraction, which converts the image to a palette mode. Then, it uses an algorithm called "nearest color" to replace the color of each pixel with the closest color on the palette. This algorithm is based on the Euclidean distance or other distance measures in color space.

b. ColorDrift: color transfer, used to convert the colors in the input image from the RGB color space to the LCH color space and calculate the difference between the target color and the template color in the LCH space. Then, the input image is transformed based on these differences to generate a new image. Based on previous migration experience, colors that do not look good after migration are filtered in the transformation space.

c. MaterialType: output of the single-label creative material recognition model, including five categories.

d. Background: single-label classification, based on transfer learning schemes. Currently, the labels for background layers include multicolor, single color, photo, and other.

e. Style: single-label classification. Currently, seven styles are supported: promotion lively, simple literary, lively cute, technological cool, avant-garde cutting-edge, business luxury, and other. The style recognition is difficult to distinguish, so we have significantly expanded the training dataset based on the manual labeling of designers to achieve good results based on the transfer learning style classification method. As shown in Fig8.

f. Category: single-label classification. The labels include real estate decoration, clothing and footwear, outdoor sports, furniture equipment, beauty care, maternity and baby products, automotive supplies, other, daily household items, food and beverage, digital appliances, bags, flowers and pets, learning and office, health care, jewelry. The category mainly divides the e-commerce categories where the creative material is placed, which are different business industry lines, and the styles designed for different industry lines have obvious differences. The res2net50 algorithm is used.

g. Scene: single-label classification. The labels include 618, Dragon Boat Festival, Children's Day, Father's Day, National Day, Labor Day, Mother's Day, other, Qixi Valentine's Day, Christmas, Taobao Double 11-12, New Year's Day, Spring Festival, Mid-Autumn Festival. The scene mainly divides the timing of the placement of the creative material, and creative materials at different times have obvious differences that match the application timing. The res2net50 algorithm is used.

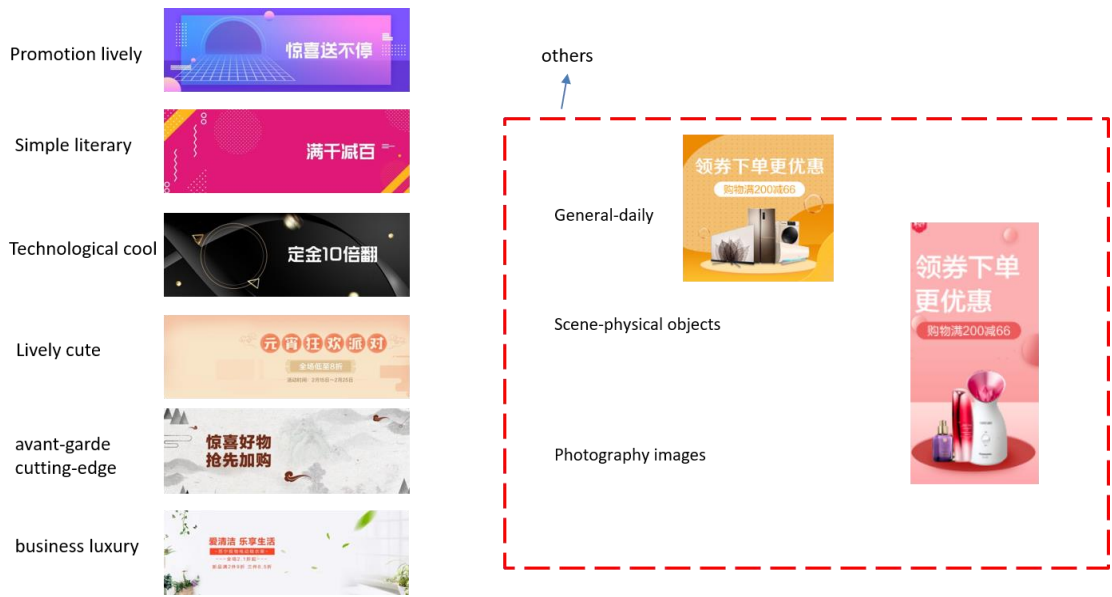

Fig 8 Display of Images in Different Styles

3.4.2 Creative Material Hierarchy

Labels The creative material hierarchy can be divided into text layers and non-text layers. Text layers include several parsing labels, which can be obtained through creative material parsing. These labels typically include font, font size, font style, etc. Non-text layers include shape layers and some special effect layers. Among them, Order/Bbox/Visible/layerType/Tone/Opacity can be parsed and obtained. Smartname is derived from the smartname module, while the remaining labels require further model parsing:

a. Color: Same as color extraction at the creative material level.

b. Color scheme: By calculating the color proportion and referring to a specific color palette, the color scheme is determined, which often represents a design style.

c. Brightness saturation: Converts the RGB image to the HSV color space and maps the brightness and saturation to three labels: vibrant, dim, and comfortable.

d. Contrast: Extracts the opacity of the primary color from the main color extraction.

e. HSV: Represents the information of color in terms of hue, saturation, and value, which better matches human visual perception. We also extract this label from the primary color.

f. HEX: Converts RGB color values to hexadecimal color values. g. Entity: Single-label classification, entity tag, including entity and non-entity. The design tone of entity and non-entity elements at the layer level of creative materials is completely different.

h. Entity_I: Single-label classification, entity subclass I. Since there are many decorative entities, we have designed multi-level entity chains, with subclass I further divided into subclass II, etc. Currently, the categories under entity subclass I include: household_goods, plants, food, decorations, animals, landscapes, and others.

i. Manifestation: Single-label classification, manifestation form. This label mainly describes the form of the material. Currently, the labels include single, multiple aggregated, and multiple scattered.

j. Style: Multi-label classification, labels include warm, minimalist, cool, romantic, technologically dazzling, lively, business luxury, cartoon, chinese retro, and general. The labels for materials are important reference for their reuse and processing.

k. Category: Multi-label classification, labels include clothing and accessories, makeups, maternity and baby, 3C digital, furniture and kitchen, appliances, festivals, automotive, food and beverages, and general.

Tab 2 Intelligent Parsing of All Label Information Table

| Creative Material Hierarchy Labels | | Creative Material Level Labels |
|---|---|---|
| Text layer | Non-text layer | |
| TextAll | | Viewbox |
| Fontname | | Layoutcount |
| Fontsize | | ColorMode |
| Bold | | Color |
| strikethrough | | Style |
| Sty | | Category |
| autoKerning | - | Scene |
| kerning | | Background |
| tracking | | Tone |
| autoLeading | | Size |
| leading | | Materialtype |
| Color | | ColorDrift |
| Order | | |
| Smartname | | |
| Opacity | | |
| Color | | |
| Bbox | | |
| Visiable | | |
| layerType | | |
| Color scheme | | |
| Tone | | - |
| Brightness saturation | | |
| Contrast | | |
| HSV | | |
| HEX | | |
| RGB | | |
| Manifestation | | |
| Entity | | |
| Entity_I | | |
| Style | | |
| Category | | |

3.3 Structured Output Layer

After obtaining the labels, the system will reconstruct and organize all creative labels and creative material hierarchy labels. The final design is to output structured data for downstream intelligent selection and intelligent creation purposes.

## 4. Experiments

### 4.1 Accuracy of Material Recognition Layer Models

In the material recognition layer, we have designed numerous recognition and detection algorithms to assist in the utilization of prior information throughout the intelligent parsing process. We have designed and analyzed the accuracy and training of these models.

#### 4.1.1 Product text area detection model

A collection of material images featuring different creative styles was gathered from Suning's internal material platform and annotated. As shown in Tab 3.

Tab3 Comparison table of accuracy for the product text area detection models.

| Model | Dets | Recall | AP | Map |
|---|---|---|---|---|
| Faster_rcnn[20] | 637 | 0.992 | 0.877 | 0.886 |
|  | 670 | 0.952 | 0.896 |  |
| Cascade_rcnn | 614 | 0.910 | 0.886 | 0.893 |
|  | 641 | 0.957 | 0.899 |  |
| Atss[21] | 730 | 0.930 | 0.881 | 0.886 |
|  | 755 | 0.941 | 0.890 |  |
| Yolov5 |  | 0.854 | 0.942 | 0.952 |

We ultimately selected the CascadeRcnn model. The recall rate of the yolov5 model is too low.

#### 4.1.2 Text area subdivision detection

We collected a batch of data from Suning's internal material platform and annotated it according to requirements. Table 4 presents the results obtained from training our model using the cascadercnn architecture.

Tab 4 Cascade Rcnn accuracy for the text area subdivision detection

| Model |  | GT | Dets | Recall | AP | Map |
|---|---|---|---|---|---|---|
| Cascadercnn | horizontal | 539 | 610 | 0.915 | 0.860 | 0.631 |
|  | Vertical | 40 | 76 | 0.8 | 0.607 |  |
|  | Mask | 174 | 207 | 0.897 | 0.791 |  |
|  | Maskbox | 48 | 75 | 0.708 | 0.625 |  |
|  | Wordart | 63 | 141 | 0.778 | 0.492 |  |
|  | shape | 20 | 38 | 0.550 | 0.412 |  |

### 4.2 Accuracy of Creative Material Recognition Models

We categorize creative materials into four types: banner, poster, product main image, detail image, and others. The training data for these categories was obtained from Suning's internal creative platform, which already includes category labels. The image types are shown in Fig 9. The accuracy of the classification model is shown in Table 5.

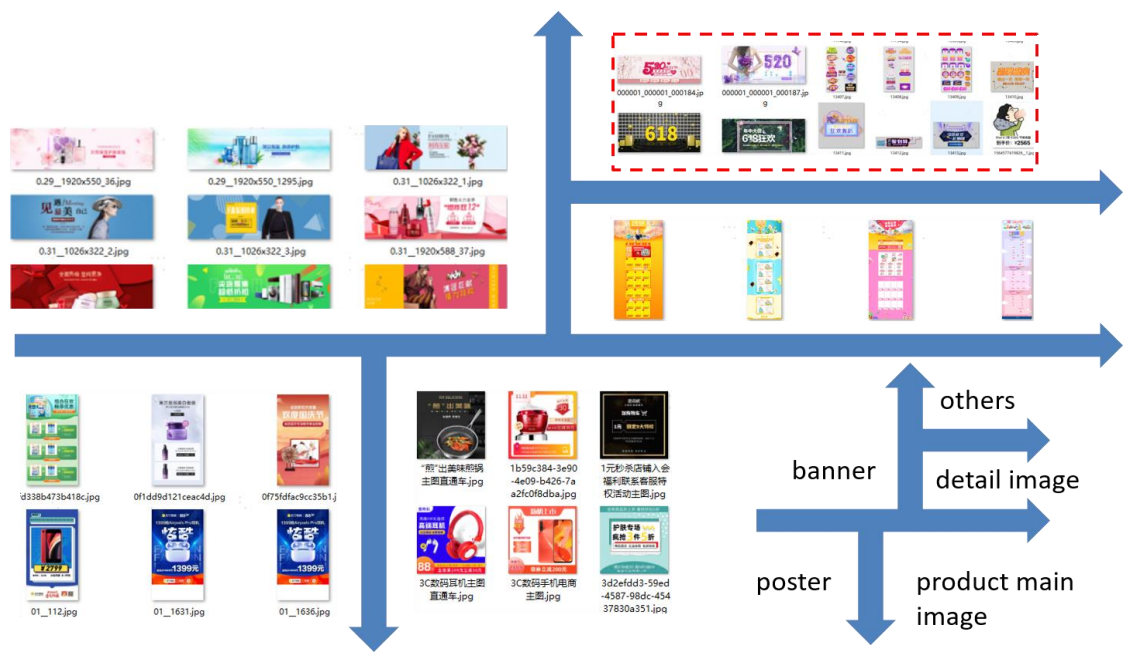

Fig 9 Examples of creative material classification.

Tab 5 Accuracy comparison table for Creative Material Recognition Models

|  | **Banner** | **Product main image** | **Poster** | **Detail image** | **Others** |
|---|---|---|---|---|---|
| **Recall** | 0.78 | 0.8 | 1 | 0.64 | 0.97 |
| **Accuracy** | 0.984 | 0.89 | 0.61 | 1 | 0.75 |

4.3 Smartname Accuracy Statistics

We tested the effectiveness of Smartname in a real workflow, sampling data from an internal design platform. A total of 110 PSD files were selected, with an average of 35 layers per PSD, for statistical analysis, as shown in Table 6.

Tab 6 Accuracy table for smartname layer identification

|  | **Goods** | **Text** | **Mc** | **Ornament** |
|---|---|---|---|---|
| Number of error-prone layers. | 50 | 27 | 59 | 28 |
| Proportion of layer errors in total layers | 0.11 | 0.05 | 0.22 | 0.12 |

We conducted statistics based on major categories, and among them, the proportion of erroneous layers in the total number of layers in that category is relatively high for MC, as the judgment for MC is comparatively more challenging.

4.4 Accuracy of Creative Material Hierarchy Labels
4.4.1 Bg label

We collected a large amount of data from Suning's e-commerce platform and assets. The schematic diagram of the corresponding label is shown in Fig 10, and the accuracy of the bg label is presented in Table 7. For the bg classification model, we ultimately chose the vggfc solution based on transfer learning.

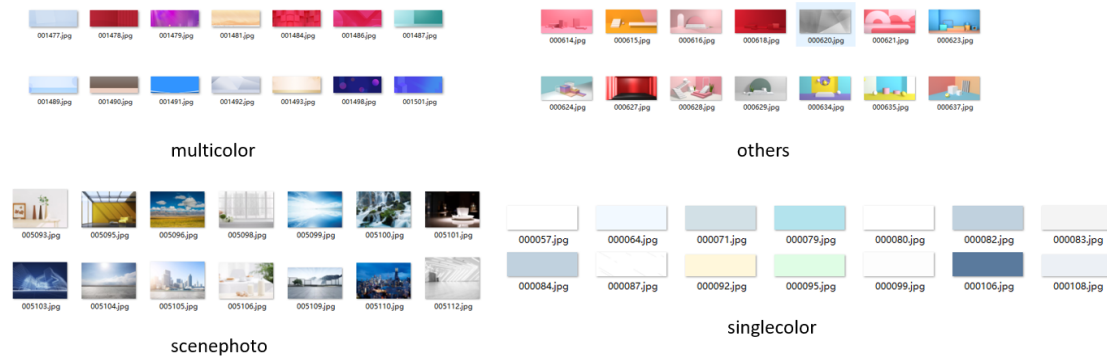

Fig 10 Illustration of the bg label

Tab 7 Accuracy table for the bg label.

| Model | | Val | Model size(m) | Time(s) |
|---|---|---|---|---|
| Vggfc | Transferlearning | 0.9246 | 1.2+57(base) | 0.061 |
| Resnet18 | Pretrained | 0.850187 | 43 | 0.062 |
| Mobilenetv2 | Pretrained | 0.8839 | 14 | 0.0557 |
| Mobilenetv3 | Pretrained | 0.850187 | 16 | 0.0538 |
| Ghostnet | No pretrained | 0.855805 | 20 | 0.0536 |

4.4.2 Style label

a. Data augmentation

The style of creative materials is difficult to label. We expand the data based on manual labeling by designers, utilizing color transfer, multi-scale cropping, and data collected from the internet. The primary rules for style expansion are as follows:

Promotion-Lively: Red, with a tendency towards dynamic and festive color schemes, often used for major promotional banners.

Simple-Literary (similar to round tables): Patterns tend to be simple, with common decorations such as flowers and plants.

Lively-Cute (featuring circular or similar characteristics): Mainly in light pink tones, often incorporating cartoonish circles, triangles, and other shapes that are not sharp, creating a gentle and cute aesthetic.

Cutting-edge-Avant-garde: Shares similarities with Minimalist - Artistic, characterized by less obvious features, yet possessing a fashionable and distinctly angular quality.

Business-Luxury: Gold, with a focus on texture.

b. Accuracy

The accuracy table for the style label is presented in Table 8. In the end, we selected the VGGFC solution based on transfer learning. The confusion matrix is shown in Fig 11. We also visualized the CAM (Class Activation Map) output of the model's feature extraction, as depicted in Fig 12, where it is evident that text regions have a significant impact on feature extraction.

Tab 8 Accuracy table for the style label

| Model | Test(%) | Model(m) | Time(s) |
|---|---|---|---|
| 3Conv+2fc | 90.15 | 10.7 | |
| Convfc_v3(3conv+fc) | 78.08 | 13 | 0.0307 |
| Convfc | 89.37 | 33 | 0.028 |
| Convfc_v2(5conv+fc) | 82.19 | 2.1 | 0.0338 |
| Resnet18 | 85.39 | 43 | 0.0364 |
| Mobilenetv2 | 85.39 | 14 | 0.047 |
| Vggfc | 94.98 | 2.1(+57) | 0.0411 |
| Resnet18_freeze | 52.96 | 43 | 0.048 |
| Vgg16_freeze | 85.84 | 57 | 0.0407 |
| Prune_resnet18 | 50.22 | 9.1 | 0.068 |
| Prune_cnn | 68.49 | 122k | 0.0489 |

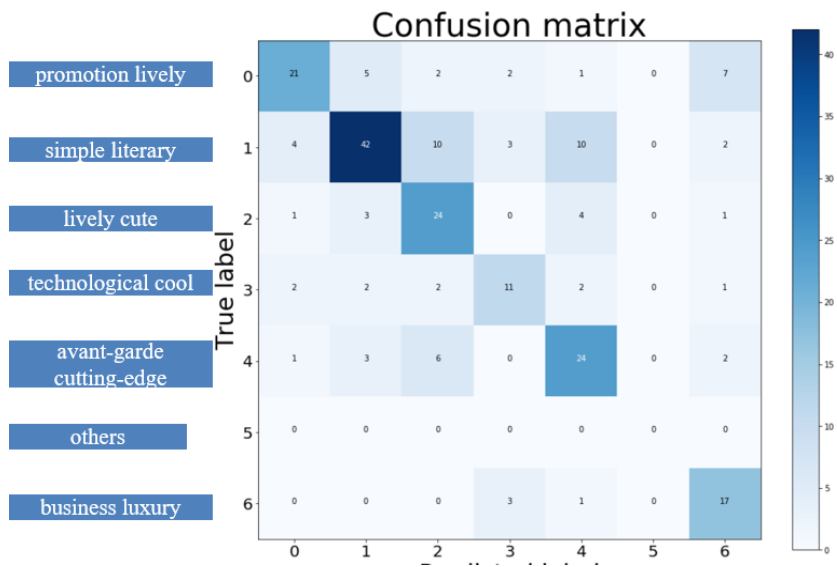

Fig 11 Confusion matrix for the style label classification model.

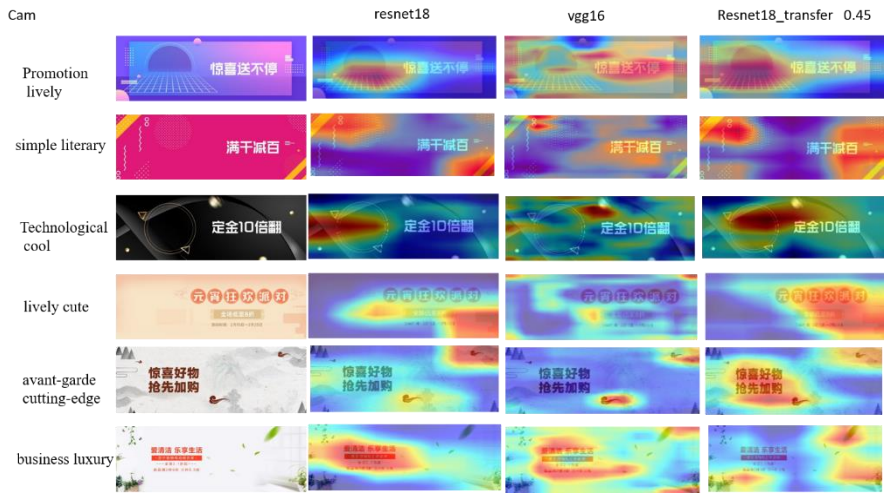

Fig 12 Cam for the style label classification model.

### 4.4.3 Category Label

a. Data Preprocessing

    We crawled some creative material data from the internet, trained a binary classifier to determine croppable data, cleaned the croppable data, trained a detector for cropping areas, and directly cropped them into usable PNG files. We annotated some of the data, trained an initial version of a single-label classifier, used this single-label classifier to annotate the already cleaned data, manually cleaned the annotation results, continued with further annotations, and continuously expanded the dataset through semi-supervised methods. In the end, we obtained a relatively large dataset.

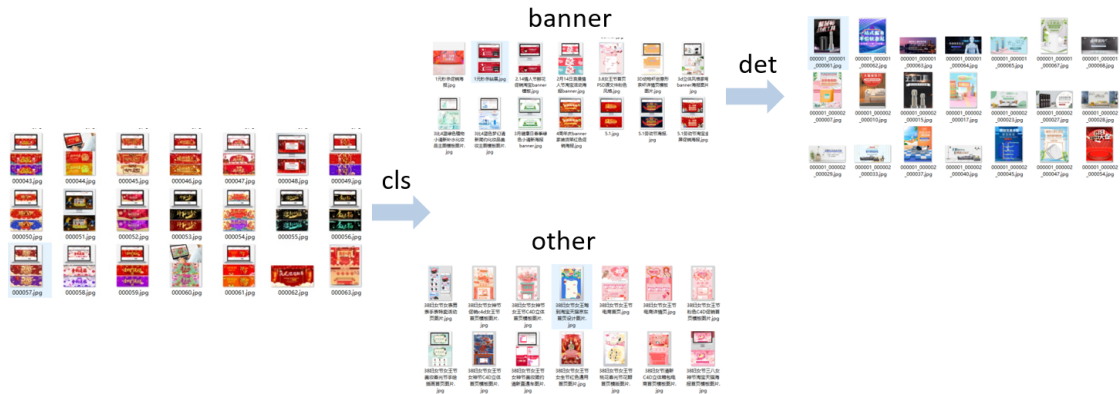

Fig 13 Illustration of data processing for Category Label.

b. Accuracy

    We ultimately chose the res2net50 model, and the accuracy comparison table is shown in Table 9

Tab 9 Accuracy table for the category label

| Model | Accuracy | Val | Model (m) | Time(s) |
| --- | --- | --- | --- | --- |
| Resnet50 | 0.71 | 0.81 | 91 | 0.0056 |
| Resnest50 | 0.76 | 0.76 | 98 | 0.0065 |
| Res2net50 | 0.67 | 0.83 | 91 | 0.0063 |
| Hrnet | 0.64 | 0.82 | 60 | 0.0059 |

| | Efficientnet | 0.72 | 0.72 | 255 | 0.014 |
| | Res2net50 | 0.74 | 0.74 | 91 | 0.072 |

#### 4.4.4 Scene Label

a. Data Preprocessing

Similar to the method of expanding data for category label.

b. Accuracy

We ultimately selected the res2net50 model, which achieved higher accuracy on the validation set.

Tab 10 Accuracy table for the Scene label

| Model | Accuracy | Best val | Model size | Time(s) |
|---|---|---|---|---|
| Res2net50 | 0.7612 | 0.8467 | 91 | 0.12 |
| Resnet50 | 0.8549 | 0.74 | 91 | 0.12 |

### 4.5 Accuracy of Hierarchical Labels for Creative Layers

For the training data of hierarchical labels for creative layers, we have annotated a large amount of original creative material data through Suning's internal annotation team. The style labels of the materials are relatively stable in terms of entity design and representation, and their forms are relatively easy to determine. However, due to the diversity of materials and the universality in design, we have adopted a multi-label approach for style and category, enriching the semantic information of the designs.

#### 4.5.1 Entity label

The entity label is a single-label classification task. The diagram of the entity labels is shown in Fig 14, and the accuracy table is presented in Table 11. In the end, we selected the single-label model based on resnet.

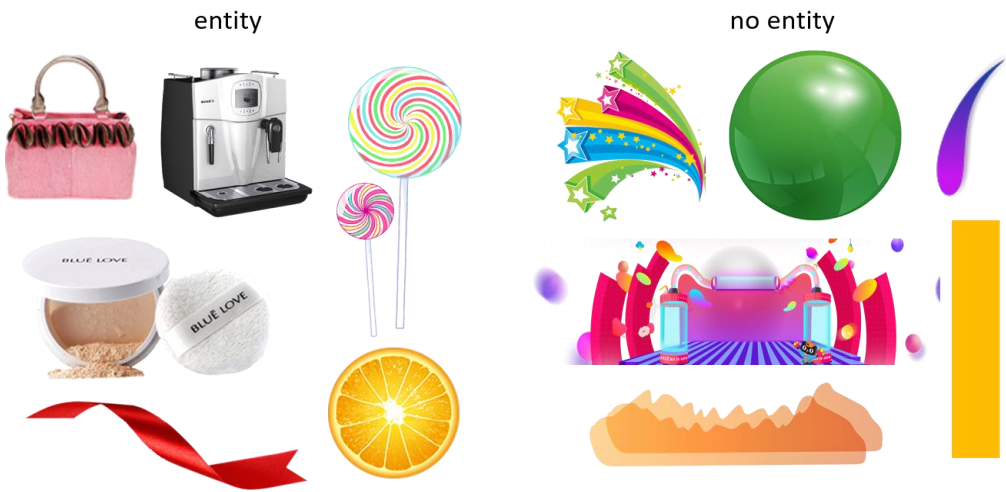

Fig 14 Examples of entity and non-entity.

Tab 11 Accuracy table for the Entity label

| Model | Accuracy | Epoch | Batch | Lr | Optimizer | Mixup | Warmup |
|---|---|---|---|---|---|---|---|

|  | Top-1 | Top-5 |  |  |  |  |  |  |
|---|---|---|---|---|---|---|---|---|
| Res2net50 | 85.0830 | 100 | 100 | 64*2 | 0.001 step | Sgd |  |  |
| Resnet | 86.5362 |  |  | 96*2 | 0.1 step | Sgd |  |  |
| Resnet | 87.8410 |  |  |  | coslr |  | √ |  |
| Resnet | 87.5148 |  |  |  |  |  | √ | √ |

### 4.5.2 Entity_I

The entity I label is a single-label classification task. The diagram of the entity I labels is shown in Fig 15, and the accuracy table is presented in Table 12. In the end, we selected the single-label model based on resnet.

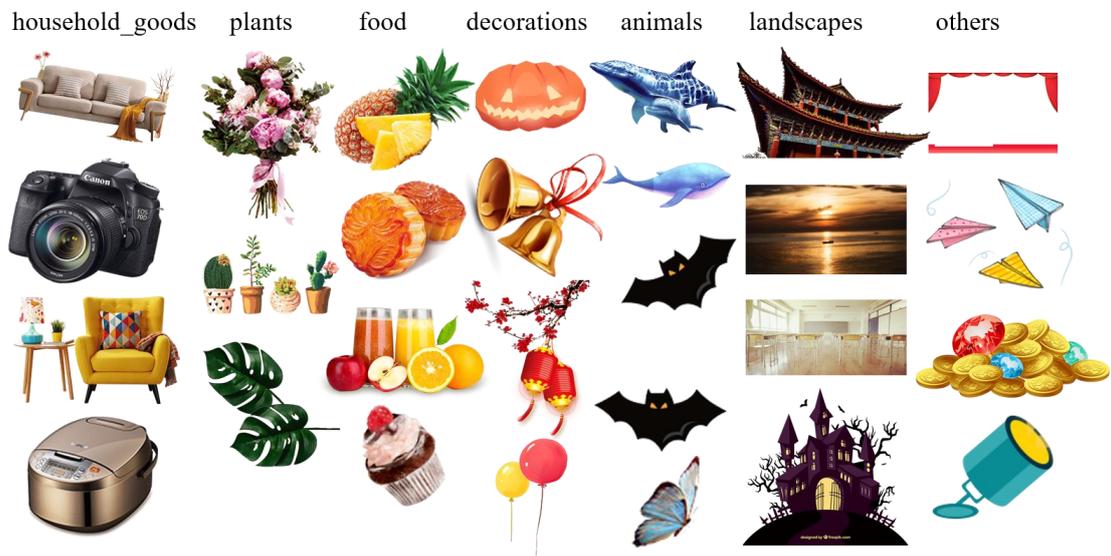

Fig 15 Examples of entity I labels

Tab 12 Accuracy table for the Entity I label

| Model | Accuracy | | Epoch | Batch | Lr | Optimizer | Label smoothing | Mixup | Cutmix | Warmup |
|---|---|---|---|---|---|---|---|---|---|---|
|  | Top-1 | Top-5 |  |  |  |  |  |  |  |  |
| Res2net50 | 52.3576 | 95.8743 | 100 | 64*2 | 0.001 step | Sgd |  |  |  |  |
| Resnetv1d | 54.2240 | 96.7583 |  | 96*2 | 0.1 step | Sgd |  |  |  |  |
| Resnet | 56.1886 | 96.2672 |  | 96*2 | 0.1 step |  |  |  |  |  |
| Resnet | 58.2515 | 96.9548 |  | 96*2 | coslr |  |  | √ |  |  |
| Resnet | 55.3045 | 95.9723 |  |  |  |  | √ |  |  |  |
| Resnet | 55.5363 | 95.9725 |  |  |  |  |  |  | √ |  |
| Resnet |  |  |  |  | 0.1 step |  |  | √ |  | √ |

### 4.5.3 Manifestation label

The Manifestation label label is a single-label classification task. The diagram of the Manifestation label labels is shown in Fig 16, and the accuracy table is presented in Table 13. In the end, we selected the single-label model based on resnet.

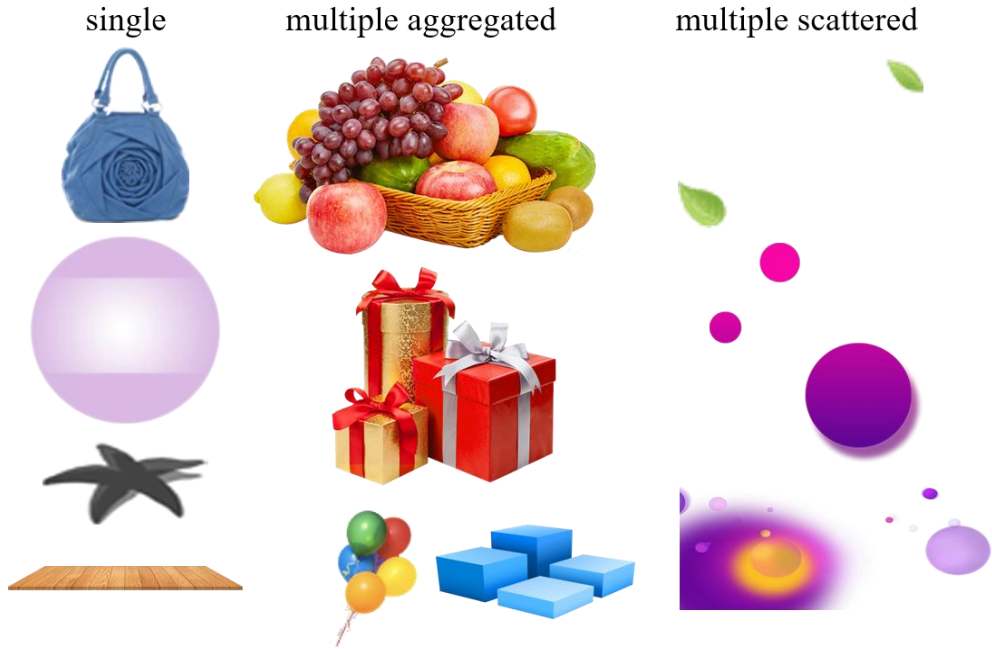

Fig 16 Examples of manifestation labels

Tab 13 Accuracy table for the manifestation labels

| Model | Accuracy | | Epoch | Batch | Lr | Optimizer | Label smoothing | Mixup | Cutmix | Pretrained |
|---|---|---|---|---|---|---|---|---|---|---|
| | Top-1 | Top-5 | | | | | | | | |
| Res2net50 | 76.5919 | 100 | 100 | 64*2 | 0.001 step | Sgd | | | | |
| Resnet | 76.2631 | 100 | 100 | 32*4 | 0.001 step | Sgd | | | | |
| Resnet | 75.9940 | 100 | 100 | 64*2 | | | | | | |
| Resnet | 76.0538 | 100 | 100 | 96*2 | | | | | | |
| Resnet | 76.8311 | 100 | 100 | 96*2 | 0.01 step | | | | | |
| Resnet_1k | 82.0329 | 100 | 100 | 96*2 | 0.1 step | | | | | |
| Resnet | 78.0867 | | | | | | | | | √ |
| Resnet | 81.0164 | | | | | | | | | |
| Resnet | 81.2855 | | | | 4ka | | | | | |
| Resnet | 81.0164 | | 100 | 96*2 | 0.1 step | | ls=0.1 | | | |
| Resnet | 81.5845 | | | | | | | a=0.2 | | |
| Resnet | 80.3587 | | | | | | | | a=1 | |
| Resnetv1d | 80.1794 | | | | | | | | | |
| Resnetv1d | 78.7145 | | | | | | | | | √ |
| Resnet | 81.7937 | | | | coslr | | | | | |

4.5.5 Style label

The creative hierarchical layers of the style label involve multi-label classification. The diagram of the multi-label categories is shown in Fig 17, and the accuracy table is presented in Table 14. We compared three multi-label models: Q2l[22], asl[11], and ml-decoder[23], and ultimately selected the ml-decoder model.

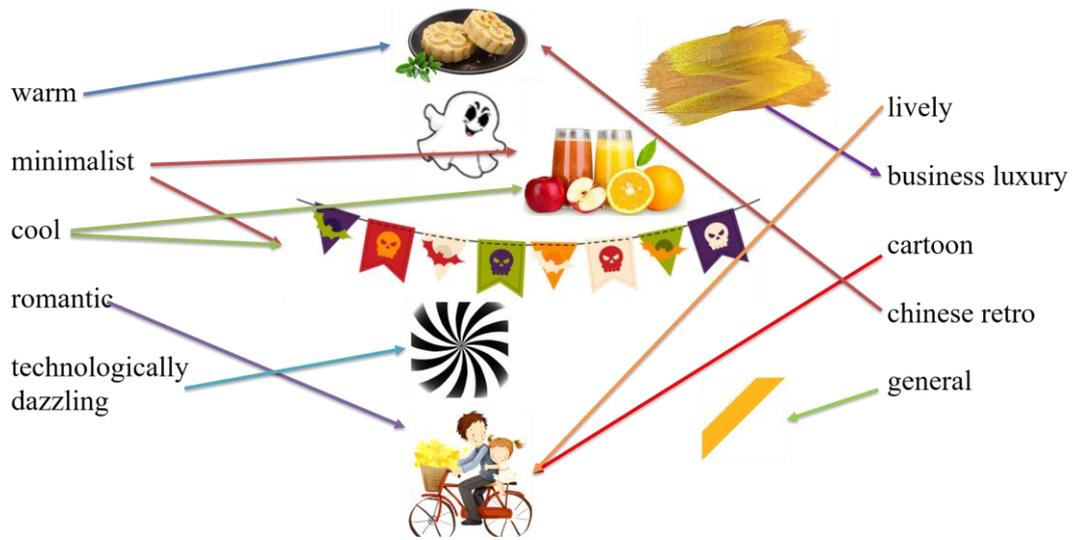

Fig 17 Examples of style milti-labels

Tab 14 Accuracy table for the style multi-labels

| Model | Backbone | Input | Map | 0.8 | | | 0.5 | | | Batch | Batch time |
|---|---|---|---|---|---|---|---|---|---|---|---|
| | | | | Precision | Recall | F1 | Precision | Recall | F1 | | |
| Asl | tresnet_m | 224 | 54.53 | 0.66 | 0.62 | 0.64 | 0.41 | 0.98 | 0.58 | 96 | 1.616/64 |
| Asl | tresnet_l | 224 | 55.08 | 0.67 | 0.60 | 0.63 | 0.40 | 0.99 | 0.57 | 46 | 1.691/64 |
| Asl | Tresnet_xl | 224 | 54.73 | 0.65 | 0.63 | 0.64 | 0.40 | 0.99 | 0.57 | 46 | 1.918/64 |
| Q2l | resnet101 | 448 | 46.82 | | | | | | | | |
| Q2l | cvt | 384 | 36.68 | | | | | | | | |
| Ml | tresnet_m | 224 | 56.78 | 0.67 | 0.63 | 0.65 | 0.43 | 0.98 | 0.59 | 36 | 2.009/64 |
| Ml | tresnet_l | 224 | 56.74 | 0.66 | 0.64 | 0.65 | 0.42 | 0.98 | 0.59 | 36 | 2.207/64 |
| Ml | tresnet_xl | 224 | 57.31 | 0.66 | 0.65 | 0.65 | 0.43 | 0.97 | 0.60 | 32 | |

4.5.6 Category label

The creative hierarchical layers of the category label involve multi-label classification. The diagram of the multi-label categories is shown in Fig 18, and the accuracy table is presented in Table 15. We also compared three multi-label models: Q2l, asl, and ml-decoder and ultimately selected the ml-decoder model.

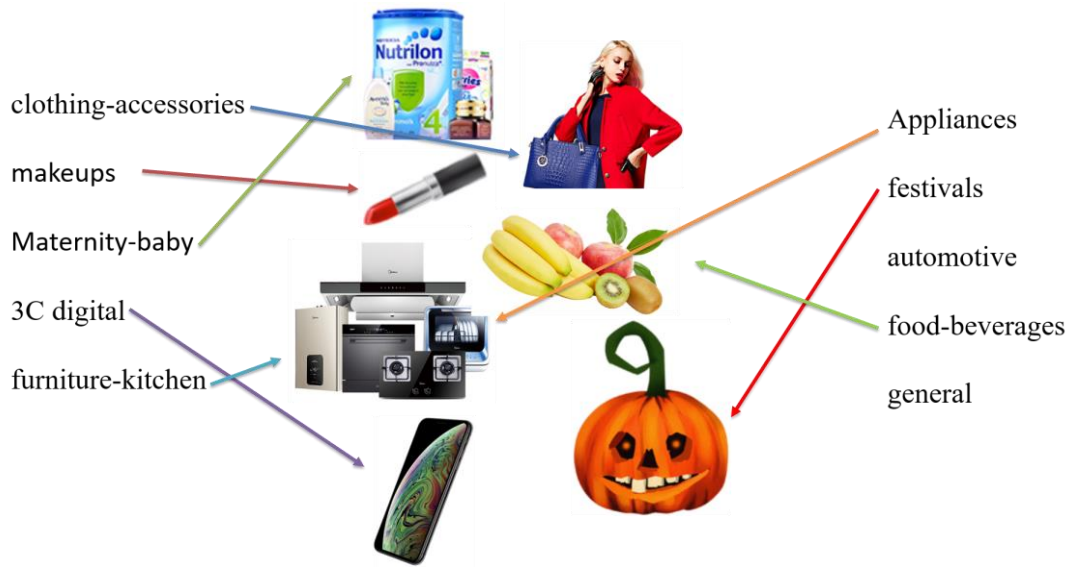

Fig 18 Examples of style milti-labels

Tab 15 Accuracy table for the style multi-labels

| Model | Backbone | Input | Map | 0.8 | | | 0.5 | | | Batch | Batch time |
|---|---|---|---|---|---|---|---|---|---|---|---|
| | | | | Precision | Recall | F1 | Precision | Recall | F1 | | |
| Asl | tresnet_m | 224 | 46.31 | 0.67 | 0.49 | 0.57 | 0.30 | 0.97 | 0.45 | 96 | 1.057/64 |
| Asl | tresnet_l | 224 | 47.32 | 0.66 | 0.54 | 0.59 | 0.30 | 0.96 | 0.46 | 96 | 1.129/64 |
| Asl | Tresnet_xl | 224 | 46.31 | 0.69 | 0.47 | 0.56 | 0.29 | 0.98 | 0.44 | 96 | 1.191/64 |
| Q2l | resnet101 | 448 | 31.46 | | | | | | | | |
| Q2l | cvt | 384 | 25.47 | | | | | | | | |
| Ml | tresnet_m | 224 | 47.93 | 0.67 | 0.52 | 0.58 | 0.33 | 0.95 | 0.49 | 36 | 1.096/64 |
| Ml | tresnet_l | 224 | 49.29 | 0.65 | 0.53 | 0.59 | 0.33 | 0.94 | 0.48 | 36 | |
| Ml | tresnet_xl | 224 | 48.13 | 0.66 | 0.53 | 0.59 | 0.31 | 0.96 | 0.47 | 32 | 1.796/64 |

5.conclusion

Intelligent parsing is one of the core components of creative design. It includes the material recognition layer, preprocessing layer, Smartname layer, and label layer. The material recognition layer encompasses various detection and recognition interfaces, covering auxiliary area detection and label recognition for both creative materials and layer hierarchies. The preprocessing layer involves filtering creative layers and grading creative materials. The Smartname layer enables intelligent naming of creative materials. The label layer provides coverage for various levels of labels for creative materials, facilitating multi-level tagging.Intelligent parsing is a comprehensive parsing framework that greatly assists downstream tasks such as intelligent creation, intelligent selection, material library construction, and the circulation and utilization of historical creatives. In Suning's actual business operations, it significantly enhances the exposure, circulation, and click-through rates of creative materials, accelerates the closed-loop production of creative materials, and generates substantial benefits.

References

1. Mackinlay J. Applying a theory of graphical presentation to the graphic design of user interfaces[C]//Proceedings of the 1st annual ACM SIGGRAPH symposium on User Interface Software. 1988: 179-189.

2. Qiang Y T, Fu Y W, Yu X, et al. Learning to generate posters of scientific papers by probabilistic graphical models[J]. Journal of Computer Science and Technology, 2019, 34: 155-169.

3. lab, A.: Intelligent code generation for design drafts. http://www.imgcook.com/ (2021)

4. Li G, Yang X. Smartbanner: intelligent banner design framework that strikes a balance between creative freedom and design rules[J]. Multimedia Tools and Applications, 2023, 82(12): 18653-18667.

5. Li G, Yang X. Two-stage dynamic creative optimization under sparse ambiguous samples for e-commerce advertising[J]. arXiv preprint arXiv:2312.01295, 2023.

6. Beltramelli T. pix2code: Generating code from a graphical user interface screenshot[C]//Proceedings of the ACM SIGCHI Symposium on Engineering Interactive Computing Systems. 2018: 1-6.

7. Xiao S, Zhou T, Chen Y, et al. UI Layers Group Detector: Grouping UI Layers via Text Fusion and Box Attention[C]//CAAI International Conference on Artificial Intelligence. Cham: Springer Nature Switzerland, 2022: 303-314.

8. Cai Z, Vasconcelos N. Cascade r-cnn: Delving into high quality object detection[C]//Proceedings of the IEEE conference on computer vision and pattern recognition. 2018: 6154-6162.

9. Li X, Wang W, Wu L, et al. Generalized focal loss: Learning qualified and distributed bounding

boxes for dense object detection[J]. Advances in Neural Information Processing Systems, 2020, 33: 21002-21012.

10. Gao S H, Cheng M M, Zhao K, et al. Res2net: A new multi-scale backbone architecture[J]. IEEE transactions on pattern analysis and machine intelligence, 2019, 43(2): 652-662.

11. Ridnik T, Ben-Baruch E, Zamir N, et al. Asymmetric loss for multi-label classification[C]//Proceedings of the IEEE/CVF International Conference on Computer Vision. 2021: 82-91.

12. Yang X, Mei T, Xu Y Q, et al. Automatic generation of visual-textual presentation layout[J]. ACM Transactions on Multimedia Computing, Communications, and Applications (TOMM), 2016, 12(2): 1-22.

13. Guo S, Jin Z, Sun F, et al. Vinci: an intelligent graphic design system for generating advertising posters[C]//Proceedings of the 2021 CHI conference on human factors in computing systems. 2021: 1-17.

14．Hua X S. The city brain: Towards real-time search for the real-world[C]//The 41st international ACM SIGIR conference on research & development in information retrieval. 2018: 1343-1344.

15. Lin T Y, Maire M, Belongie S, et al. Microsoft coco: Common objects in context[C]//Computer Vision–ECCV 2014: 13th European Conference, Zurich, Switzerland, September 6-12, 2014, Proceedings, Part V 13. Springer International Publishing, 2014: 740-755.

16. Robinson A. Sketch2code: Generating a website from a paper mockup[J]. arXiv preprint arXiv:1905.13750, 2019.

17. Liu Z, Chen C, Wang J, et al. Owl eyes: Spotting ui display issues via visual understanding[C]//Proceedings of the 35th IEEE/ACM International Conference on Automated Software Engineering. 2020: 398-409.

18. Chen J, Chen C, Xing Z, et al. Unblind your apps: Predicting natural-language labels for mobile gui components by deep learning[C]//Proceedings of the ACM/IEEE 42nd International Conference on Software Engineering. 2020: 322-334.

19. Su T. FSMdroid: guided GUI testing of android apps[C]//Proceedings of the 38th International Conference on Software Engineering Companion. 2016: 689-691.

20. Girshick R. Fast r-cnn[C]//Proceedings of the IEEE international conference on computer vision. 2015: 1440-1448.

21. Zhang S, Chi C, Yao Y, et al. Bridging the gap between anchor-based and anchor-free detection via adaptive training sample selection[C]//Proceedings of the IEEE/CVF conference on computer vision and pattern recognition. 2020: 9759-9768.

22. Liu S, Zhang L, Yang X, et al. Query2label: A simple transformer way to multi-label classification[J]. arXiv preprint arXiv:2107.10834, 2021.

23. Ridnik T, Sharir G, Ben-Cohen A, et al. Ml-decoder: Scalable and versatile classification head[C]//Proceedings of the IEEE/CVF Winter Conference on Applications of Computer Vision. 2023: 32-41.